\documentclass{article}

\usepackage{microtype}
\usepackage{graphicx}
\usepackage{subfigure}
\usepackage{booktabs} 
\usepackage{siunitx}
\usepackage{url}
\usepackage{hyperref}

\usepackage[accepted]{icml2021}

\icmltitlerunning{Stimulating student engagement with an AI board game tournament}

\begin{document}

\twocolumn[

\icmltitle{Stimulating student engagement with an AI board game tournament}




\begin{icmlauthorlist}
\icmlauthor{Ken Hasselmann}{ecam,ulb}
\icmlauthor{Quentin Lurkin}{ecam}
\end{icmlauthorlist}

\icmlaffiliation{ecam}{ECAM Brussels School of engineering, Brussels, Belgium}
\icmlaffiliation{ulb}{Université libre de Bruxelles, Brussels, Belgium}

\icmlcorrespondingauthor{Ken Hasselmann}{hsl@ecam.be}
\icmlcorrespondingauthor{Quentin Lurkin}{lur@ecam.be}

\icmlkeywords{gamification, teaching AI, ICML}

\vskip 0.3in
]



\printAffiliationsAndNotice{}  

\begin{abstract}
Strong foundations in basic AI techniques are key to understanding more advanced concepts.
We believe that introducing AI techniques, such as search methods, early in higher education
helps create a deeper understanding of the concepts seen later in more advanced AI and algorithms courses.
We present a project-based and competition-based bachelor course that gives second-year 
students an introduction to search methods applied to board games.
In groups of two, students have to use network programming and AI methods to build an AI agent to 
compete in a board game tournament---{\it othello} was this year's game. 
Students are evaluated based on the quality of their projects and on their performance
during the final tournament.
We believe that the introduction of gamification, in the form of competition-based learning,
allows for a better learning experience for the students.

\end{abstract}

\section{Introduction}
\label{introduction}

Recent years have seen increased interest in machine learning and deep learning: many courses 
and teaching material are now available~\cite{wollowski_survey_2016,ye_artificial_2021}, and their availability is growing.
We believe that, regardless of the importance of teaching the most recent and state-of-the-art technologies,
it is important for students to have sound knowledge of data structures and the basics of AI algorithms.
In the case of this course, when we refer to the basics of AI algorithms, we refer to search algorithms, adversarial 
search algorithms, and the basics of game theory~\cite{russell_artificial_2021}.

We designed a project-based programming course for second-year students revolving around the creation of an AI
agent for a given board game.
Students are presented with a board game at the beginning of the course. In groups of two, they have to use 
the knowledge gained throughout the lectures to create an AI agent that will compete against all other 
student groups' agents by the end of the course period.

The objectives of this course are three-fold: 
(i) we want to introduce students to different python programming paradigms that they have not seen in the curriculum yet;
(ii) we want to introduce AI for games and game theory; and
(iii) we want to introduce source code management and unit testing for their project.

We focus on practical work, group projects, and creating a challenge by having students compete with each other.
The introduction of gamification, namely competition-based learning, in teaching has shown positive results 
in student motivation and engagement~\cite{seaborn_gamification_2015, burguillo_using_2010}. We therefore believe that gamifying the course's end-goal, by introducing a student tournament, will improve the overall
learning experience. 
Students have to apply the lecture concepts immediately and are motivated by the idea
of defeating rival teams, creating some intrinsic motivation for learning.

\section{Course design}

This section details the course design, i.e., the structure and organisation, both for lectures and practical sessions.

\subsection{Audience}

This course is designed for second-year engineering students enrolled in a three-year bachelor's program, 
which is usually followed by a two-year master's degree program.
It is taught to students that chose to have a pre-specialization in computer science
and electronics. It is one of the courses that gives students a grasp of the computer science specialization
and therefore plays an important role in the students' decisions when later selecting a specialization subject for their studies.

\subsection{Lecture topics}
\label{sec:topics}

Being a course intended both as an introduction to some specific programming paradigms in python and
as an introduction to AI for games, the lecture topics are divided into three main parts.

The first part of the lectures focuses on programming paradigms. The lectures cover mainly:
(i) network programming, where students learn about the basics 
of networking and protocols in order to exchange messages on a local network programmatically; and
(ii) concurrent programming, where they learn about threads and how to handle multiple concurrent computations.

The second part of the lectures focuses on classical AI for board games, namely: 
(i) basic data structures, such as stacks, heaps, and trees;
(ii) search algorithms, such as breadth-first search, depth-first search, The Dijkstra algorithm, 
and A\* search; and
(iii) adversarial search algorithms, such as the minimax algorithm,
negamax, alpha-beta pruning, and iterative deepening.

The third part of the course presents the basic concepts of source code management with the use of git.
In this part, the main concepts of git and its most common commands are presented, and students also
learn about the importance of unit testing, how to implement good unit tests, and the concept of code coverage.


\subsection{Project}

\begin{figure}[tb]
  \centering
  \includegraphics[scale=0.18]{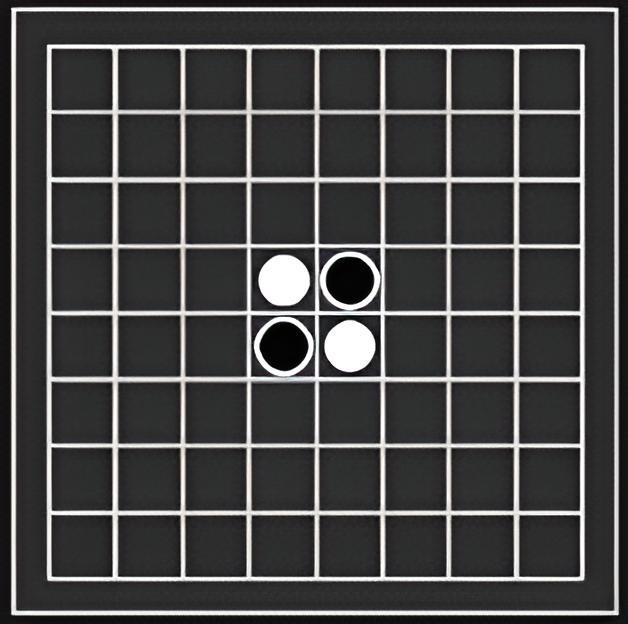}
  \caption{\textbf{The board of othello}. Othello is a two-player strategy game (a fixed initial setup variant of reversi),
  where players take turns placing disks of their assigned color on the board. Disks have one black and one white side. When placing a disk, any of the opponent's disks that are in line and bounded by disks of the current player are turned over to the current player's color. Once the board is full, at the end of the game, the player whose color is assigned to a majority of disks wins the game.}
  \label{fig:othello}
\end{figure}

In order to motivate students to grasp all the concepts described in the previous section, we ask them to build
an AI agent for a board game.
The board game to be tackled in the project changes every year. This year, the board game was {\it othello}---see Fig.\ref{fig:othello}.

We ask the students to form groups of two. The project then consists in coming up with a good strategy to play
the game, implementing it, and letting it compete in the final tournament at the end of the semester.
We provide the students with a simple description of how to connect and use the 
tournament system (see Section~\ref{sec:gamingsystem} for a full description
of this system).

We observed that typically students first take some time understanding the rules of the game, then starting to prototype
strategies on paper, before they start to code.
We ask them to code their AI agents in python. We do not provide any specific instructions on what algorithm
to use. We want students to experiment using different data structures and AI techniques to try and find a suitable one.

\subsection{Tournament system}
\label{sec:gamingsystem}

The tournament system works as follows: a game server hosts the game and agents (clients) connect to this
server to start a game and make a move on the board.
The game server works by accepting direct TCP connection from agents with a documented custom protocol.

Students have to build agents to connect to this server and play the game.
Automatic matchmaking is implemented into the game server so that every client that is connected plays
a game against all other clients that are connected.
The game server waits for connections from agents. Once at least two agents are connected, the matchmaking starts
and the first game is launched. The game server saves the state of the current game being played and queues all other games
to be played.
Once a game is started, an agent has \SI{5}{s} to make a move at each turn, if the agent does not send an instruction within these
\SI{5}{s}, or if the move made by the agent is illegal, the game server registers a bad move, 
and the opposing agent takes a turn.
At the end of the game, the full log of the game is saved, including bad moves and the winning agent.

\begin{figure}[tb]
  \centering
  \includegraphics[scale=0.17]{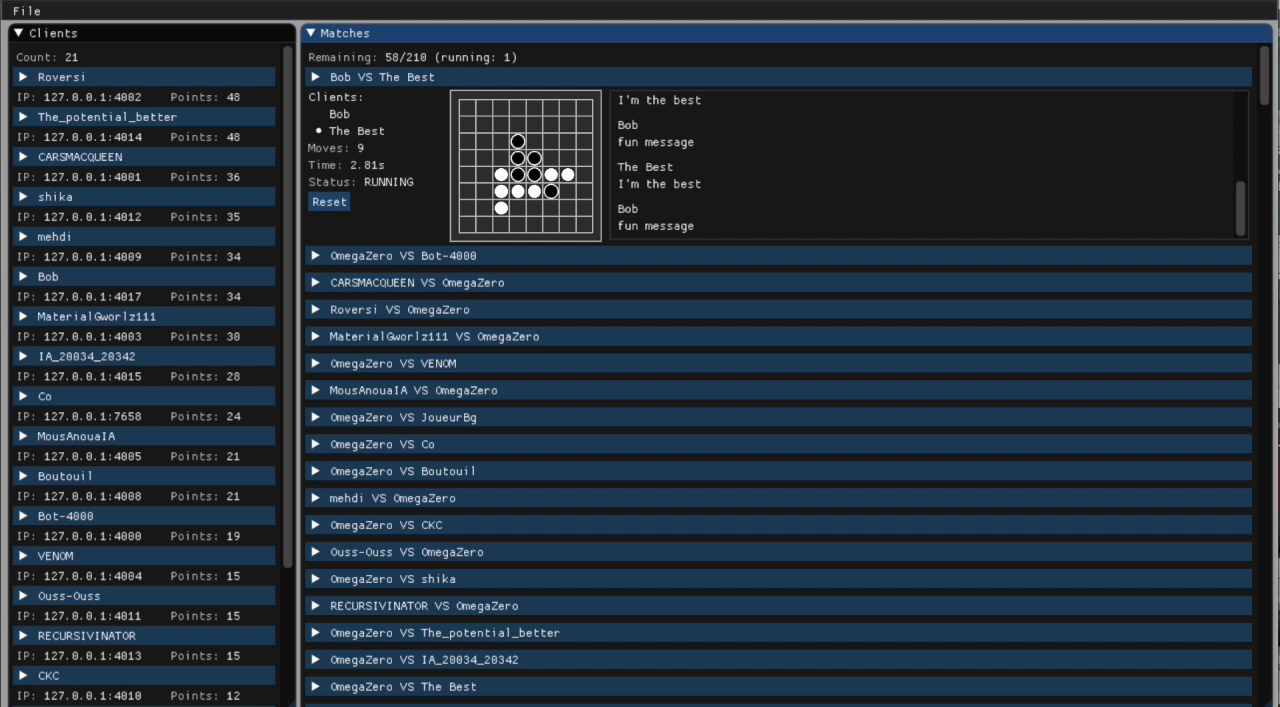}
  \caption{\textbf{GUI of the tournament software}. The current game being played is displayed on the top right; below, is the list of queued games. The left column shows all connected clients (AI agents).  }
  \label{fig:competsoft}
\end{figure}

During practical sessions, the game server runs on a teacher's computer so that students can test their agents
in the local network of the classroom. 
A random agent---an agent that plays a random move on the game board every turn---is constantly connected 
to the game server during the practical sessions so that 
any student agent connecting to the server automatically competes against it. This allows students to 
test their agents in a simple scenario.
All games being played on the server are displayed on the server's GUI (see Fig. \ref{fig:competsoft}).

We also provide the students with the code of the game server. This allows them to
review and inspect code written by an external person, and helps them in their own implementation. 
They can also run the game server locally on their own computer to, for example, test their agents at home.
We wish students to carefully read the description of the game server, 
understand how the system works, and propose an implementation for their agent 
to connect and use the custom network protocol.

The playing system is thus designed to enforce that students use all notions seen in the lectures,
if they want to succeed in connecting and playing the game.

\subsection{AI agents}

For the creation of the AI agent, we do not provide any boilerplate implementation,
nor do we provide the implementation of the random agent.
Students have access to the descriptions of classical algorithms presented during
the lectures~(see section~\ref{sec:topics}) and, of course, to any external resources they could find.
We do not enforce any specifications on the algorithms to be used.
We want students to experiment with different data structures 
and AI techniques to try and find a suitable one.

A typical implementation of an agent includes: 
(i) a network API for communicating with the game server;
(ii) a model of the game's rules to determine possible moves given a game board;
(iii) an algorithm for determining the next move to be played.

The students do have complete freedom on the AI algorithm to use for determining the next move.
Our minimal expectation for this part is that student model the game using a tree of possible actions and
implement a search algorithm for navigating the tree to find a move.
We choose games with a relatively high branching factor on purpose so that exhaustive search is impractical.
A highly effective implementation might, for instance, employ Monte-Carlo tree search 
with a custom-trained heuristic.

\subsection{Evaluation}
\label{sec:eval}

Students are evaluated on different aspects and at different stages during the course.
The first part of the evaluation consists of individual programming exercises done in class, during a one hour exam period.
Students complete four programming exercises with precise specifications.
For this exam, they only have access to a computer, the lecture materials, and the latest version of 
the python documentation.
The programs produced by the students are then individually tested.
The students' scores depend on whether their code passes all unit tests.

The second part of the evaluation is on the creation of the board game AI agent.
Each group of students has to submit their work using an online git repository. 
Repositories are then fetched and all agents from all students compete against each other
in an online tournament.
In this part, students are evaluated on a competition among AI agents from all student groups,
but also on the quality of their code, the quality of their unit tests, the coverage of those tests, and
how they managed their source code in git.
Since the proposed board game is different every year, the difficulty of the task of creating the AI agent
can vary slightly. However, we attempt to minimize this by selecting board games with a similar degree 
of complexity, or by qualitatively adjusting our level of expectation for the project.

Each of the two parts (the individual programming exercise and the group project) accounts for 50\% for the final course score.
In this way, we evaluate the individual students' understanding of the lecture topics as well as the quality
of their group projects, in terms of source code quality, code efficiency, and source code management.

\subsection{Gamification components}

Motivation is a key factor in student performance and engagement in a course~\cite{saeed_how_2012}.
Several studies report that the use of gamification is an efficient way of motivating students~\cite{seaborn_gamification_2015,shahid_review_2019,jawad_gamifying_2021}.
We chose to base our course on competition-based learning, a certain way of implementing gamification
in which students take part in a friendly competition.
Competition-based learning has been shown to improve motivation and 
help increase student performance~\cite{burguillo_using_2010,ho_impact_2022}. 
Several studies describe its usage in computer science courses~\cite{lawrence_teaching_2004,ebner_successful_2007,ribeiro_teaching_2009,jung_case_2010}
and report satisfactory results in student engagement and motivation. 
It can also be used in conjunction with project-based learning
to encourage peer-learning among students~\cite{boss_reinventing_2007,burguillo_using_2010,ho_impact_2022}.
As described in Section~\ref{sec:eval}, we evaluate students on multiple criteria. 
%
We designed the evaluation score awarded to the project to be in several parts: 35\% of the score is 
awarded based on the use of git, code coverage, and software documentation; another 35\% of the score
is awarded based on software engineering practices, naming conventions, comments, and bare minimum functionalities;
the remaining 30\% depends on the students' performance during the final tournament.
The performance of the students in the tournament (the aforementioned 30\%) is computed based on:
(i) their agent ranking better than the random agent;
(ii) the number of bad moves that their agent tried to play;
(iii) the overall rank of the agent.
Since the project score accounts for 50\% of the final course grade, the tournament performance
only accounts for 15\% of the final grade of the student.

There are two frameworks for introducing competition in education.
In competitive-based learning, the learning result depends on the result of the competition itself.
In competition-based learning, by contrast, the learning result is independent of the student's score 
in the competition~\cite{burguillo_using_2010}.
In our case, the different parts introduced in the grading system make the 85\% of the final score independent
from the competition. 
The learning result is thus also independent from the competition since, the course's skills are diverse,
and a good final score can be achieved with a well-designed but relatively simple AI agent.

It has been shown that is some cases, competitive goal structures can impede achievement, and
that cooperation, or cooperation with intergroup competition, is more effective than interpersonal competition~\cite{johnson_effects_1981}.
That is why we designed the project as a group work, and also why we decided on 
the relatively small percentage devoted to the final competition:
to encourage so-called friendly competition, where students are motivated
to engage in the competition seriously without jeopardizing peer-based learning.

\section{Course assessment}

We did not yet perform a systematic and formal teaching assessment campaign. We base the judgements reported here on
the informal interactions we had with students and the overall atmosphere during the lectures and practical sessions.

Overall, our experience with the students has been very positive. Although it was not mandatory to attend, 
there was high attendance during both lectures and practical sessions, compared to other courses that also did not have
mandatory attendance.

Most student groups seemed to show interest in understanding how the algorithms they saw during the lectures 
were working and came up with their own implementations of the algorithms seen in the lectures.

All student groups managed to create a working agent that interacted with the game server using the proper protocol.
Less than 15\% of student groups' agents tried to play bad moves of the game.
Around 75\% of student groups managed to create agents that were able to beat the
random agent in some games. Some student groups' agents would still lose when the random agent was the first player. 
Note that playing first indeed gives an advantage in the game.
The top 15\% of student groups experimented with different heuristics to estimate the 
quality of a board position in adversarial search algorithms.
The best group implemented an iterative deepening minimax algorithm with alpha-beta pruning 
and transposition tables.

Some students reported that the course motivated them to continue their study in the field of computer science.
As this course is given quite early in the studies, it is for us very important that it is designed to be
attractive to students.

\section{Conclusion}

We have presented our design for an introductory course to AI and software engineering. The course involves group project-based learning
and gamification through competition-based learning.
The results of the course suggest that both project-based learning and gamification help motivate students and create a better
learning experience through peer-based learning and the intrinsic motivation created by friendly competition.

We wish to conduct a formal teaching assessment campaign before next year's edition of the course to confirm what 
our informal assessment has suggested about the students' highly positive sentiment towards this course.

\section*{Software and Data}

The tournament system software is available online~\cite{lurkin_tournament_2021} (documentation in french).



\section*{Acknowledgements}

We thank the reviewers for their careful reading of our manuscript and their
insightful comments and suggestions.
We acknowledge the support of ECAM Brussels School of engineering.




\bibliography{teachingml}
\bibliographystyle{icml2021}


\end{document}